\documentclass[accepted]{uai2021} 
\usepackage[american]{babel}

\usepackage{amsmath,amsfonts,bm}

\def\eqref#1{equation~\ref{#1}}

\def\1{\bm{1}}

\DeclareMathAlphabet{\mathsfit}{\encodingdefault}{\sfdefault}{m}{sl}
\SetMathAlphabet{\mathsfit}{bold}{\encodingdefault}{\sfdefault}{bx}{n}

\newcommand{\E}{\mathbb{E}}

\newcommand{\Dval}{\pazocal{D}'}
\newcommand{\D}{\pazocal{D}}

\DeclareMathOperator*{\argmax}{arg\,max}
\DeclareMathOperator*{\argmin}{arg\,min}

\usepackage{hyperref}
\usepackage{url}
\usepackage{microtype}
\usepackage{graphicx}
\usepackage{subfigure}
\usepackage{booktabs} 
\usepackage{calrsfs}
\usepackage{amsmath}
\usepackage{multirow}
\usepackage{natbib}
\usepackage{amssymb}
\usepackage{mathtools}
\usepackage{wrapfig}
\usepackage{gensymb}
\usepackage{amssymb}
\usepackage{pifont}
\newcommand{\cmark}{\ding{51}}%
\newcommand{\xmark}{\ding{55}}%

\usepackage[american]{babel}

\usepackage{natbib} 
    \renewcommand{\bibsection}{\subsubsection*{References}}
\usepackage{mathtools} 
\usepackage{booktabs} 
\usepackage{tikz} 

\DeclareMathAlphabet{\pazocal}{OMS}{zplm}{m}{n}

\usepackage{algpseudocode}
\usepackage{algorithm}

\usepackage[capitalize]{cleveref}
\Crefname{lm}{Lemma}{Lemmas}
\Crefname{prop}{Proposition}{Propositions}
\Crefname{defn}{Definition}{Definitions}
\Crefname{thm}{Theorem}{Theorems}
\Crefname{assumption}{Assumption}{Assumptions}
\Crefname{cor}{Corollary}{Corollaries}
\Crefname{condition}{Condition}{Conditions}

\title{Post-hoc loss-calibration for Bayesian neural networks}

\author[1]{\href{mailto:Meet P. Vadera <mvadera@cs.umass.edu>?Subject=Your UAI 2021 paper}{Meet P. Vadera \thanks{Work partially done during internship at IBM Research.}}{}}
\author[2]{Soumya Ghosh}
\author[2]{Kenney Ng}
\author[1]{Benjamin M. Marlin}
\affil[1]{%
    College of Information and Computer Sciences\\
    University of Massachusetts Amherst\\
    Amherst, MA, USA
}
\affil[2]{%
    IBM Research\\
    Cambridge, MA, USA
}

\begin{document}

\maketitle

\begin{abstract}
Bayesian decision theory provides an elegant framework for acting optimally under uncertainty when tractable posterior distributions are available. Modern Bayesian models, however, typically involve intractable posteriors that are approximated with, potentially crude, surrogates. This difficulty has engendered loss-calibrated techniques that aim to learn posterior approximations that favor high-utility decisions. In this paper, focusing on Bayesian neural networks, we develop methods for correcting approximate posterior predictive distributions encouraging them to prefer high-utility decisions. In contrast to previous work, our approach is agnostic to the choice of the approximate inference algorithm, allows for efficient test time decision making through amortization, and empirically produces higher quality decisions. We demonstrate the effectiveness of our approach through controlled experiments spanning a diversity of tasks and datasets.  
\end{abstract} 

\section{Introduction}

    
Decision-making under uncertainty is a frequently encountered challenge across diverse applications. Examples include, medical diagnosis \citep{Leibig2017LeveragingUI}, hazard alarms \citep{Economou2016OnTU}, and autonomous driving \citep{McAllister2017ConcretePF}. Bayesian decision theory (BDT) provides an elegant framework for decision-making under uncertainty. Given the Bayes posterior, a set of decisions, and a utility function encoding user preferences, BDT dictates that the decision that maximizes the expected utility, where the expectation is with respect to the Bayes posterior, is optimal. While promising, it is worth noting that the optimality guarantees provided by BDT only hold when the true Bayes posterior is available.    

Inspired in part by the success of deep learning, modern Bayesian models are often high dimensional, not restricted to conjugate prior-likelihood families, and almost always have intractable posterior distributions. Bayesian neural networks (BNN), the primary focus of this paper, are a prime example of such models. In BNNs, lacking tractable posterior distributions, various, at times crude surrogates are employed to approximate the posterior. Decisions that maximize the expected utility, with the expectation computed with respect to the surrogate rather than the true posterior, are not guaranteed to be optimal. This observation has engendered research into loss-calibrated inference~\citep{lacoste2011approximate, Cobb2018LossCalibratedAI,  kusmierczyk2019variational} techniques that modify the approximate inference procedures to prefer regions of the posterior, most relevant to the decision-making task at hand.  These methods, however, intricately couple posterior inference with decision-making. Such coupling poses several difficulties. Any change in the utility function necessitates recomputing the loss-calibrated posterior. Approximate inference algorithms are typically computationally expensive, and such re-computations can be computationally onerous.
Moreover, updating the posterior on account of a modified utility function rather than updated prior beliefs or data is conceptually unappealing. Furthermore, under the loss-calibrated inference framework, every posterior inference algorithm requires a bespoke loss-calibrated counterpart to be developed. Such loss-calibrated variants may be challenging to develop, and no obvious counterpart may exist for a practitioner's algorithm of choice. 

Inspired by these difficulties, we focus on post-hoc corrections to the posterior predictive distributions. By choosing to correct predictions, we are able to decouple posterior inference from the process of correcting decisions. As a result, our proposed method is agnostic to the choice of the inference algorithm; any off-the-shelf procedure may be used. Indeed, given a posterior approximation and an \emph{unlabeled} calibration dataset, we do not even need to access the original training data. Together these properties significantly expand the applicability of loss-calibrated inference approaches.
Further, we use a single neural network to parameterize the posterior predictive corrections. At test time, decision-making involves a single forward pass through the network. This provides significant speed-ups over alternate loss-calibrated approaches~\citep{Cobb2018LossCalibratedAI} which require expensive Monte Carlo approximations. We empirically demonstrate that our approach can support applications that require real-time decisions.  We also find that the post-hoc corrections can be efficiently learned and, since they do not involve any posterior inference inexpensively adapted to changing utility functions. Finally, through careful experiments across diverse applications, we demonstrate that the aforementioned conceptual and computational benefits do not come at the expense of empirical performance. Our post-hoc correction procedure performs as well or better than competing approaches.

\section{Related Work}
\paragraph{Bayesian Neural Networks}
 Let $p(y|\mathbf{x}, \theta)$ represent the probability distribution induced by a deep neural network classifier over classes $y\in\pazocal{Y}=\{1,..,C\}$ given feature vectors $\mathbf{x}\in\mathbb{R}^D$. Given training data $\pazocal{D} =\{(\mathbf{x}_i,y_i)|1\leq i\leq N\}$ they are commonly learned through maximum conditional likelihood maximization. Instead of attempting to find the single (locally) optimal set of parameters $\theta_*$, Bayesian neural networks are learned by inferring a posterior distribution $p(\theta \mid \pazocal{D},\theta^0)$ over the unknown parameters $\theta$ given the prior $p(\theta \mid \theta^0)$ with hyper-parameters $\theta^0$. Predictions on unseen data points $\mathbf{x}_*$ are made via the \emph{posterior predictive distribution} which involves averaging over the uncertainty in the posterior distribution,
\begin{align}
\allowdisplaybreaks[4]
\label{eq:posterior_predictive}
p(y_*\mid \mathbf{x}_*, \pazocal{D},\theta^0) &= \int p(y_*\mid \mathbf{x}_*, \theta) p(\theta\mid\pazocal{D},\theta^0) d\theta .
\end{align}
 
Applying Bayesian inference to neural networks is challenging because both the posterior and the posterior-predictive distributions are intractable to compute, and require approximations. We now briefly review various approximate inference algorithms that have been used to approximate the intractable posterior, including variational inference (VI) \citep{jordan1999introduction} and Markov Chain Monte Carlo (MCMC) methods \citep{Neal:1996:BLN:525544,welling2011bayesian}.

In VI, an auxiliary distribution $q_{\phi} (\theta)$  is defined to approximate the true parameter posterior $p(\theta | \pazocal{D},\theta^0)$. The variational parameters $\phi$ are selected to minimize the Kullback-Leibler (KL) divergence between $q_{\phi} (\theta)$ and $p(\theta \mid \pazocal{D},\theta^0)$. \citet{hinton1993keeping} did early work applying VI to neural networks. \citet{graves2011practical} and \citet{blundell2015weight} later developed stochastic variants of VI that scale to modern networks. In other related work, authors \citet{hernandez2015probabilistic, soudry2014expectation, ghosh2016assumed} explored assumed density filtering (ADF) and expectation propagation \citep{minka2001expectation} based approaches \citep{li2015stochastic, hernandez2016black} for learning BNNs.
%
%
These approaches however result in biased posterior estimates for complex posterior distributions. MCMC methods on the other hand provide sampling-based posterior approximations that are unbiased, but are often computationally more expensive to use. MCMC methods allow for drawing a correlated sequence of samples $\theta_t \sim p(\theta|\pazocal{D},\theta^0)$ from the parameter posterior. The samples can then be used to approximate the posterior predictive distribution as a Monte Carlo average as shown in \eqref{eq:MC_posterior_predictive},
\begin{align}
\allowdisplaybreaks[4]
\label{eq:MC_posterior_predictive}
p(y| \mathbf{x}, \pazocal{D},\theta^0) 
\approx  \frac{1}{T}\sum_{t=1}^T p(y|\mathbf{x}, \theta_t), \quad 
\theta_t &\sim p(\theta|\pazocal{D},\theta^0).
\end{align}
While Hamiltonian Monte Carlo \citep{Neal:1996:BLN:525544} remains the gold standard for inference in BNNs, its stochastic gradient variants \citep{welling2011bayesian, chen2014stochastic} are popular for large networks, and we will use them extensively in this paper.
\paragraph{Bayesian decision theory and loss-calibrated inference}
Bayesian decision theory provides a framework for decision-making under uncertainty \citep{Berger1988StatisticalDT}. Under the framework, we elicit a utility function $\mathbf{u}(h, y)$, where $h$ denotes a decision within a set of possible actions $\pazocal{A}$ and $y$ denotes model predictions. Next, given a data point $\mathbf{x_*}$, we evaluate the expected utility (also known as the \emph{conditional gain}), $\pazocal{G}(h \mid \mathbf{x_*})$, for all $h \in \pazocal{A}$ using the utility function $u(.)$ and the posterior predictive distribution $p(y_*| \mathbf{x_*}, \theta, \pazocal{D})$,
\begin{align}
\label{eq:predictive_conditional_gain}
&\pazocal{G}(h  \mid \mathbf{x_*}) = \int_{y_*} u(h, y_*)p(y_* \mid \mathbf{x_*}, \pazocal{D}, \theta^0) d\mathbf{y_*} 
\end{align}
Finally, we select the optimal decision $c_*$, such that it maximizes the conditional gain, $c_* = \argmax_{h \in \pazocal{A}}\pazocal{G}(h \mid \mathbf{x_*})$.
However, an important assumption in this frameworks is that we have access to the \emph{true} posterior predictive distribution. As noted earlier, the true posterior predictive distribution is intractable for Bayesian neural networks. Rather, in practice, we only have access to a crude approximation $\tilde{p}(y_* \mid \mathbf{x}_*, \pazocal{D}, \theta^0)$. Using this approximation as an drop-in replacement to $p(y_* \mid \mathbf{x}_*, \pazocal{D}, \theta^0)$ in \cref{eq:predictive_conditional_gain} no longer guarantees optimality of decisions, $c_*$. 

This observation has inspired research in loss-calibrated inference. \citet{lacoste2011approximate} presents a variational approach for Gaussian process classification that derives from lower-bounding the log-conditional gain. To train the variational distribution, it presents an EM algorithm with closed form updates which alternates between sampling from the variational posterior and making optimal decisions under the variational posterior. \cite{Cobb2018LossCalibratedAI} extends the work done by \citet{lacoste2011approximate} to Bayesian neural networks, and derives an objective that is a cost-penalized version of the standard evidence lower-bound (ELBO). Both \citet{lacoste2011approximate} and \citet{Cobb2018LossCalibratedAI} deal with only discrete values for the decisions $h$, \citet{kusmierczyk2019variational} generalizes these methods to continuous decisions. Beyond variational approximations, \citet{abbasnejad2015loss} present  an importance sampling-based approach that encourages high utility decisions. 

Finally, this paper was in part inspired by the work of \citet{kusmierczyk2019correcting}. Similarly to us, they propose corrections to model predictions instead of the posterior approximations. However, unlike us, their primary focus is on problems with low-dimensional posteriors. As a result, their methods are challenging to apply to the large BNN models considered here. 



\section{Post-hoc corrections for posterior predictive distributions}
Going forward, we will assume that we have access to a calibration dataset $\Dval \coloneqq \{\mathbf{x}_n\}_{n=1}^N$ and that we can evaluate the posterior predictive distribution, under some approximation to the posterior, at all $\mathbf{x}_n \in \Dval$. The log conditional gain on $\Dval$ is,
\begin{equation}
\begin{split}
\log \;  &\pazocal{G}(\mathbf{h} = \mathbf{c} \mid \Dval) = \\ & \sum_{n=1}^{N} \log \int_{y} u(h=c_n, y_n=y)p(y_n=y \mid \mathbf{x}_n, \pazocal{D}, \theta^0)d{y},
\end{split}
\label{eq:log_conditional_gain}
\end{equation}
where $\mathbf{c} = \{c_n\}_{n=1}^N$, and $c_n = \argmax_{h \in \pazocal{A}}\pazocal{G}(h \mid \mathbf{x_n})$.

If we had access to the \emph{true} posterior predictive distribution, guarantees from Bayesian decision theory ensure that the decisions $c_n$ are optimal. However, for BNNs we only have access to potentially crude approximations to the posterior and $c_n$ are no longer guaranteed to be optimal. To address this, we introduce an utility aware correction, $q(y_n \mid \mathbf{x}_n, \lambda)$ to the (approximate) posterior predictive distribution $p(y_n \mid \mathbf{x}_n, \D, \theta^0)$ evaluated at $\mathbf{x}_n \in \Dval$. The correction is parameterized by a set of learnable parameters,  $\lambda$. In our experiments, we use a neural network to parameterize $q$ and $\lambda$ corresponds to the weights of that network. We observe that the log conditional gain can be expressed as a function of $\lambda$,
\begin{equation}
\small
\begin{split}
    &\log \; \pazocal{G}(\mathbf{h} = \mathbf{c} \mid \Dval; \lambda) = \\ &\sum_{n=1}^{N} \log \E_{q(y_n=y | \mathbf{x}_n, \lambda)} \left[\frac{p(y_n=y | \mathbf{x_n}, \pazocal{D}, \theta^0) u(h=c_n, y_n=y)}{q(y_n=y | \mathbf{x}_n, \lambda)} \right],
\end{split}
\end{equation}
\normalsize
and is lower bounded by,  
\begin{equation}
\begin{split}
\pazocal{U}(\lambda, \mathbf{c}, \Dval) = &\sum_{n=1}^{N} \E_{q(y_n | \mathbf{x}_n, \lambda)}\left[ \log u(c_n, y_n) \right]
\\ & - \textrm{KL}\big[q(y_n | \mathbf{x}_n, \lambda) || p(y_n | \mathbf{x}_n, \pazocal{D}, \theta^0)\big], \\
\end{split}
\label{eq:lobj}
\end{equation}
where the bound $\log \; \pazocal{G}(\mathbf{h} = \mathbf{c} \mid \Dval; \lambda) \geq \pazocal{U}(\lambda, \mathbf{c} ; \Dval)$ follows from Jensen's inequality. See ~\cref{app:derivation} for a detailed derivation. 
%
We learn $q(\cdot \mid \cdot, \lambda)$ by maximizing $\pazocal{U}(\lambda, \mathbf{c} ; \Dval)$ with respect to $\lambda$ and $\mathbf{c}$. Our algorithm proceeds in an coordinate ascent fashion by alternating between fixing $\mathbf{c}$ and taking a gradient step in the direction maximizing $\pazocal{U}(\lambda, \mathbf{c} ; \Dval)$ with respect to $\lambda$ and then fixing $\lambda$ and maximizing $\mathbf{c}$. We limit our attention to finite discrete-valued decision problems prevalent in classification settings. For these problems, we are able to trivially maximize $\mathbf{c}$ given $\lambda$ by enumerating the expected utility of all decisions and selecting the highest utility decision. 

\paragraph{The variational gap} between the log conditional gain and the lower bound,  
\begin{equation}
\small
\begin{split}
    \displaystyle \log \; &\pazocal{G}(\mathbf{h} = \mathbf{c} \mid \Dval; \lambda) - \pazocal{U}(\lambda, \mathbf{c}, \Dval) \\&= \sum_{\mathbf{x}_n \in \pazocal{D}'} \textrm{KL}\big[q(y_n | \mathbf{x}_n, \lambda) || \frac{p(y_n | \mathbf{x}_n, \pazocal{D}, \theta^0)u(c_n, y_n)}{Z_n} \big],
\end{split}
\label{eq:vgap}
\end{equation}
\normalsize
where $Z_n = \E_{p(y_n | \mathbf{x}_n, \pazocal{D}, \theta^0)} [u(c_n, y_n)]$, lends further insights into the optimization problem. For a fixed $\mathbf{c}$, maximizing \cref{eq:lobj} is equivalent to minimizing the KL divergence between  $q$ and the original posterior predictive distribution scaled by the utility function, pointwise over the calibration dataset. This further highlights a key aspect of the proposed approach, it corrects the (\emph{typically}) low-dimensional posterior predictive distribution rather than the unwieldy, high-dimensional BNN posterior. The lower bound \cref{eq:lobj} also lends itself to an intuitive interpretation. The first term guides $q(\cdot \mid \cdot, \lambda)$ to higher utility decisions while the second  Kullback-Leibler divergence term encourages $q(\cdot \mid \cdot, \lambda)$ to be close to the approximate posterior predictive distribution in the KL sense.

Although nearly operational, two key challenges remain in applying the developed framework. The first stems from computational considerations necessary when working with large Bayesian models like BNNs.  Posterior predictive distributions for BNNs need to be approximated via Monte Carlo simulations. Computation and storage cost of Monte Carlo approximations grow linearly with the number of samples and can be prohibitive for large networks. The other challenge stems from  user preferences typically being expressed as cost functions~\citep{Berger1988StatisticalDT, kusmierczyk2019variational} rather than utility functions, and yet our development thus far has dealt exclusively with utility functions. We next describe strategies effective at alleviating both these concerns.

\subsection{Practical considerations}
\paragraph{Amortized posterior predictive distribution} We tackle the computational concerns associated with Monte Carlo approximations to the posterior predictive distribution by learning an amortized approximation \citep{balan2015bayesian, Vadera2020GeneralizedBP}. We use the online distillation algorithm proposed by \cite{balan2015bayesian}, a special case of the general framework of \citet{Vadera2020GeneralizedBP},  and distill the posterior predictive distribution into a single ``student'' neural network model. This algorithm aims to minimize the Kullback-Leibler (KL) divergence between $p(y_n \mid \mathbf{x}_n, \pazocal{D}, \theta^0)$ and a student network $S(y_n \mid \mathbf{x}_n, \omega)$, parameterized by $\omega$ for $\mathbf{x}_n \in \Dval$. The online nature of this algorithm allows us to amortize the computation of posterior predictive distribution, without having to instantiate a large number of posterior samples. Once we have trained the student model, we can use it as a drop-in replacement for the posterior predictive distribution in \cref{eq:lobj},
\begin{equation}
\begin{split}
\pazocal{U}^s(\lambda, \mathbf{c}, \Dval) = 
&\sum_{n=1}^{N} \E_{q(y_n | \mathbf{x_n}, \lambda)}\left[ \log u(c_n, y_n) \right] 
\\ &- \textrm{KL}(q(y_n \mid \mathbf{x_n}, \lambda) || S(y_n \mid \mathbf{x_n}, \omega)).
\end{split}
\label{eq:post_hoc_correction_objective}
\end{equation}
\textbf{Decision cost v/s utilities: } In practical applications it is common to have user preferences encoded as decision costs rather than utilities. We follow \cite{kusmierczyk2019variational},to translate between costs and utilities. 
Let us denote the decision cost function as $\ell(h, y)$, where $h$ again denotes the decision and $y$ denotes the predicted class. W can re-write the utility function as  $u(h, y) = M - \ell(h, y)$, where $M \geq \sup_{h, y} \ell(h, y)$. By substituting this in \cref{eq:lobj} we obtain, 
\begin{equation}
\begin{split}
\pazocal{L}(\lambda, \mathbf{c}; \Dval) = &\sum_{n=1}^{N} \E_{q(y_n | \mathbf{x_n}, \lambda)}\left[ \log \left(M - \ell(c_n, y) \right) \right]\\ & - \textrm{KL}(q(y_n | \mathbf{x_n}, \lambda) || p(y_n | \mathbf{x_n}, \pazocal{D}, \theta^0)),
\end{split}
\label{eq:lcost}
\end{equation}
and the analogous amortized variant is given by, 
\begin{equation}
\begin{split}
\pazocal{L}^s(\lambda, \mathbf{c}; \Dval) = &\sum_{n=1}^{N} \E_{q(y_n | \mathbf{x_n}, \lambda)}\left[ \log \left(M - \ell(c_n, y) \right) \right]\\
&- \textrm{KL}(q(y_n | \mathbf{x_n}, \lambda) || S(y_n | \mathbf{x_n}, \omega)).
\end{split}
\label{eq:lcost_student}
\end{equation}
Further performing a first order Taylor series expansion about $M$ \citep{kusmierczyk2019variational, lacoste2011approximate} we obtain,
\begin{equation}
\small
\begin{split}
\pazocal{L}^s(\lambda, \mathbf{c}; \Dval) \approx \sum_{n=1}^{N} \E_{q(y_n | \mathbf{x_n}, \lambda)}\left[ \log M  - \frac{\ell(c_n, y)}{M} \right] &\\ 
 - \textrm{KL}(q(y_n | \mathbf{x_n}, \lambda) || S(y_n | \mathbf{x_n}, \omega)),
\end{split}
\label{eq:lcost_student_taylor_step_1}
\end{equation}
Noting that $\E_{q(y_n | \mathbf{x_n}, \lambda)} [\log M]$ is constant with respect to $\lambda$ and $\mathbf{c}$, we arrive at, 
\begin{equation}
\begin{split}
\pazocal{\tilde{L}}^s(\lambda, \mathbf{c}; \Dval) = - &\sum_{n=1}^{N} \E_{q(y_n | \mathbf{x_n}, \lambda)}\left[ \frac{\ell(c_n, y)}{M}   \right] \\ 
& - \textrm{KL}(q(y_n | \mathbf{x_n}, \lambda) || S(y_n | \mathbf{x_n}, \omega)).
\end{split}
\label{eq:lcost_student_final}
\end{equation}
If we have access to the original posterior predictive distribution $p(y_n | \mathbf{x_n}, \pazocal{D}, \theta^0))$, the analogous objective is,
\begin{equation}
\begin{split}
\pazocal{\tilde{L}}(\lambda, \mathbf{c}; \Dval) = - &\sum_{n=1}^{N} \E_{q(y_n | \mathbf{x_n}, \lambda)}\left[ \frac{\ell(c_n, y)}{M}   \right] \\ 
& - \textrm{KL}(q(y_n | \mathbf{x_n}, \lambda) || p(y_n | \mathbf{x_n}, \pazocal{D}, \theta^0))) .
\end{split}
\label{eq:lcost_final}
\end{equation}
Our experiments maximize either $\pazocal{\tilde{L}}^s(\lambda, \mathbf{c}; \Dval)$ or $\pazocal{\tilde{L}}(\lambda, \mathbf{c}; \Dval)$ depending on the experimental setup. An algorithm giving an overview of our approach is presented in \cref{app:algorithm}.

With the description of our method complete, we reemphasize the distinct advantages provided by it. Observe that we only require that we are either able to evaluate the posterior predictive distribution or an amortized approximation to it on $\Dval$. We remain agnostic and make no assumptions about how the posterior or the posterior predictive distributions were computed. Moreover learning the corrections, $q(\cdot \mid \cdot, \lambda)$, involves optimizing \cref{eq:lcost_student_final} or \cref{eq:lcost_final} and is no more expensive than training standard deep neural networks. Finally, at a test point $\mathbf{x}_*$ the expected cost associated with a decision $h$ is $\sum_{k=1}^C\ell(h, y=k)q(y=k \mid x_*, \lambda)$. Computing this expected cost involves a \emph{single} forward pass through $q$. Our framework, thus, amortizes test time decision-making. This leads to significant speed-ups over existing loss-calibrated inference approaches, which must first compute the posterior predictive distribution by performing an expensive Monte Carlo integration over the corrected posterior before making decisions. Test time amortization allows our method to be used in applications that demand real-time decision-making.  We summarize our approach's similarities and differences to relevant work in \cref{tab:1}.

\begin{table}[htbp]
\small
\centering
\caption{\textbf{Overview of related loss-calibrated methods}. \emph{Inference Agnostic}: Method does not modify or make assumptions about the posterior inference algorithm. \emph{Scalable}: Method scales to modern Bayesian neural networks learned from large data. \emph{Amortized Decisions}: Method does not require multiple forward passes for test time decision-making.}
\resizebox{\linewidth}{!}{
\begin{tabular}{cccccc}
\toprule
 & \begin{tabular}[c]{@{}c@{}}Inference \\ Agnostic \end{tabular} & \begin{tabular}[c]{@{}c@{}}Scalable \end{tabular} & \begin{tabular}[c]{@{}c@{}}Amortized \\ Decisions\end{tabular}\\ \midrule
\cite{lacoste2011approximate}  &\xmark  &\xmark  &\xmark \\ \midrule
\cite{Cobb2018LossCalibratedAI}     & \xmark  &   \cmark & \xmark\\ \midrule
\cite{kusmierczyk2019variational} & \xmark &  \cmark & \xmark \\ \midrule
\cite{kusmierczyk2019correcting} & \cmark & \xmark & \cmark\\ \midrule
Ours & \cmark & \cmark  & \cmark  \\
\bottomrule
\end{tabular}%
}
\label{tab:1}
\end{table}
\normalsize

\section{Experiments}
In this section, we carefully vet our proposed method against relevant baselines across a diverse range of applications. 
Broadly, we divide our experiments into three major categories to target three practical scenarios --- 1) Decision-making in poor data quality regimes, 2) Decision-making with a reject option (also known as \emph{selective classification}), 3) Decision-making under real-time constraints. 

Throughout this section, we experiment with fully-connected and ResNet18 \citep{krizhevsky2012imagenet} architectures. We test our methods on data from MNIST~\citep{lecun1998mnist}, CIFAR10~\citep{krizhevsky2009learning}, and the challenging CamVid~\citep{BrostowSFC:ECCV08} dataset. We demonstrate that our posterior correction consistently improves the quality of decisions when used in conjunction with popular BNN inference algorithms --- black-box variational inference (BBVI)~\citep{blundell2015weight}, stochastic gradient Hamiltonian Monte-Carlo (SGHMC)~\citep{chen2014stochastic}, and Kronecker-factored Laplace approximation (KFAC-Laplace)~\citep{ritter2018scalable}. Noting that SGHMC~\citep{Yao2019QualityOU} typically provides a more faithful approximation to the BNN posterior, we restrict ourselves to SGHMC for real data experiments. To compare against a loss-calibrated inference procedure, we develop a loss-calibrated variant of SGHMC. Following \citet{lacoste2011approximate} we define the following utility scaled posterior,
\begin{equation}
    \Tilde{p}(\theta| \pazocal{D}, u) \propto p(\theta| \pazocal{D}) \pazocal{G(\mathbf{h^*} | \pazocal{D})}.
\end{equation}
The loss-calibrated stochastic gradient HMC (LC-SGHMC) algorithm then proceeds by sampling from this scaled posterior using SGHMC. Given that SGHMC is typically more accurate than competing variational methods, we view LC-SGHMC as a strong loss-calibrated inference baseline. We also provide a brief overview of SGHMC in \cref{app:SGHMC}.


\subsection{Synthetic Data Experiments}
We begin with experiments on synthetic data, employing fully connected architectures and three popular inference techniques, BBVI with local reparameterizations~\citep{kingma2015variational}, SGHMC, and KFAC-Laplace. 

\begin{figure}
    \centering
    \includegraphics[width=0.95\linewidth]{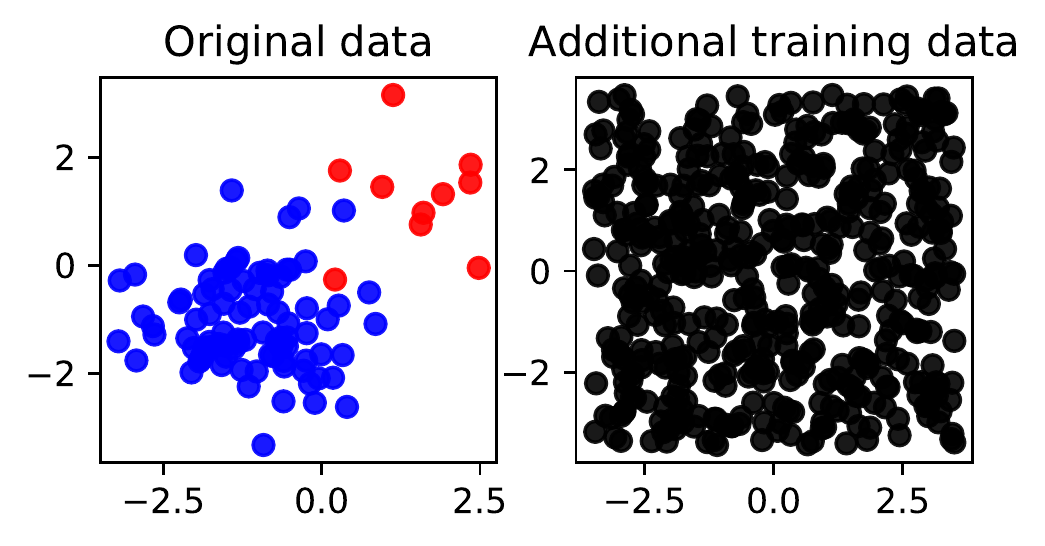}
    \includegraphics[width=0.95\linewidth]{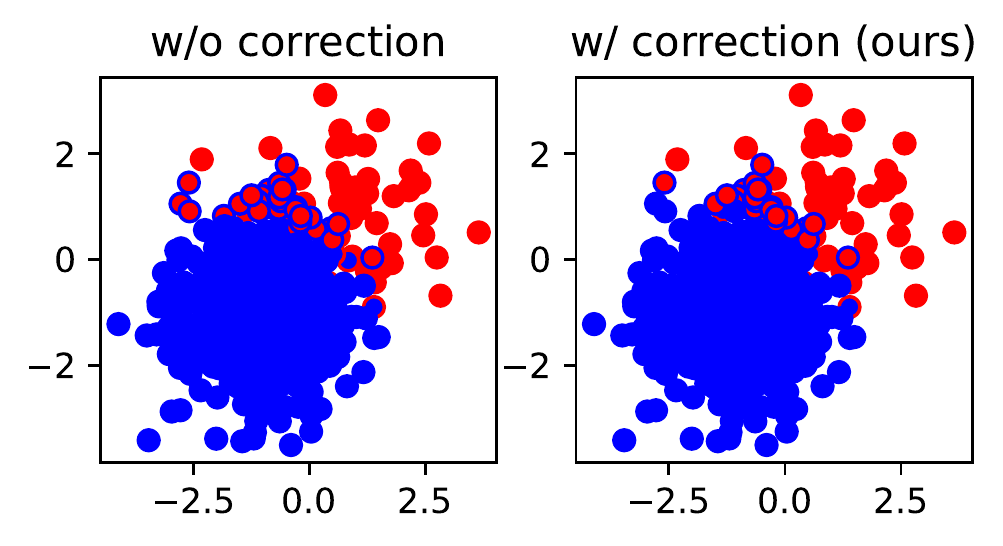}
    \caption{\textbf{Synthetic Data (Top)}. Labeled training data ($\pazocal{D}$) is shown in the left plot and unlabeled calibration data ($\pazocal{D'}$) is shown in the right plot. Blue markers represent the negative class and red markers represent the positive class. \textbf{Decision Visualization (Bottom)}. A visual comparison of decision-making using BBVI without correction, and with our correctional approach for a single trial, on test data drawn from the same distribution as the training data $\pazocal{D'}$. The edge colors indicate ground truth classes, while the face colors indicate the predicted classes. Consistency between edge and face colors indicate correct predictions.}
    \label{fig:2d_data}
\end{figure}

\textbf{Experimental setup:} We construct a two-dimensional, two class data set with class imbalance. We generate data from the two classes by sampling isotropic Gaussian distributions with means $[-1, -1]$ and $[+1, +1]$. For training, we use $90$ data instances from the negative class and $10$ data instances from the positive class. We resample the Gaussian distributions to create a test set, which again contains $90$ negative examples and $10$ positive ones. For calibration data, we uniformly sample the two dimensional space to generate $500$ unlabeled data instances. We repeat this procedure ten times and generate ten training and calibration datasets. \cref{fig:2d_data} visualizes one of these ten datasets. For each dataset, we learn a $50$ unit, single hidden layer, multi-layer perceptron with ReLU activations using BBVI, SGHMC, and KFAC-Laplace and use a $100$ sample Monte Carlo approximation to compute the corresponding posterior predictive distributions. We employ the following decision-cost function,
\begin{align*}
    \ell(c, y) = \left\{ \begin{array}{ll}
        0, & \text{for } y=c \\
        1, & \text{for } y \neq c, y= \text{positive} \\
        0.1, & \text{for } y \neq c, y= \text{negative} 
        \end{array} \right.,
\end{align*}
which encourages decisions that minimize false negative errors for the minority class --- often a desirable property in practice. In this experiment, we learn the corrections by maximizing \cref{eq:lcost_final}. Additional experimental details can be found in \cref{app:synthetic_data_experiments}. 

\paragraph{Results: } We compute the average decision cost under our cost-function for each of the three inference algorithms with and without our post-hoc correction on the test set. 
Table \ref{tab:toy_data_results} summarizes our results, where the error bars stem from having repeated the experiment on the ten randomly generated training and calibration datasets. Our post-hoc correction results in test decisions with  lower decision costs compared to the decisions produced by the uncorrected variants. The costs are only marginally lower in this synthetic example, where the approximations to the posterior are likely already good.  In the following, we will see that the decision costs can be substantially lower in more challenging scenarios. In subsequent experiments, we solely rely on stochastic gradient HMC algorithms for approximating the Bayesian neural network posterior. A full panel of results is presented in Appendix \ref{app:selective_decision_masking}.
\begin{table}[htbp]
\centering
\caption{\textbf{Results on synthetic data}. Test decision costs with and without post-hoc correction over $10$ replicates. Post-hoc correction consistently provides lower cost decisions. Results presented as mean $\pm$ std. dev.}
\label{tab:toy_data_results}
\resizebox{.9\linewidth}{!}{
\begin{tabular}{ccc}
\toprule
 & \begin{tabular}[c]{@{}c@{}}W/O post-hoc \\ correction\end{tabular} & \begin{tabular}[c]{@{}c@{}}W/ post-hoc \\ correction (ours)\end{tabular}  \\ \midrule
VI       & 0.019 $\pm$ 0.011 & 0.016 $\pm$ 0.010 \\ \midrule
SGHMC    & 0.018 $\pm$ 0.008 &  0.017 $\pm$ 0.009  \\ \midrule
KFAC-Laplace & 0.021 $\pm$ 0.007 & 0.018 $\pm$ 0.008   \\
\bottomrule
\end{tabular}%
}
\end{table}

\subsection{Selective Classification}
\begin{figure*}
    \centering
    {\includegraphics[width=0.245\textwidth]{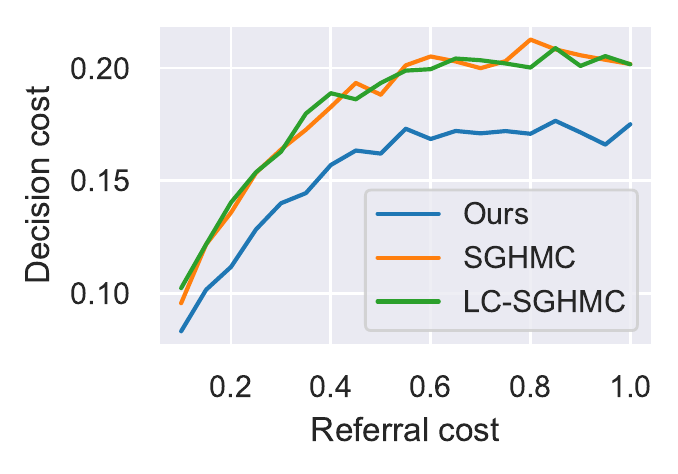}}
    {\includegraphics[width=0.245\textwidth]{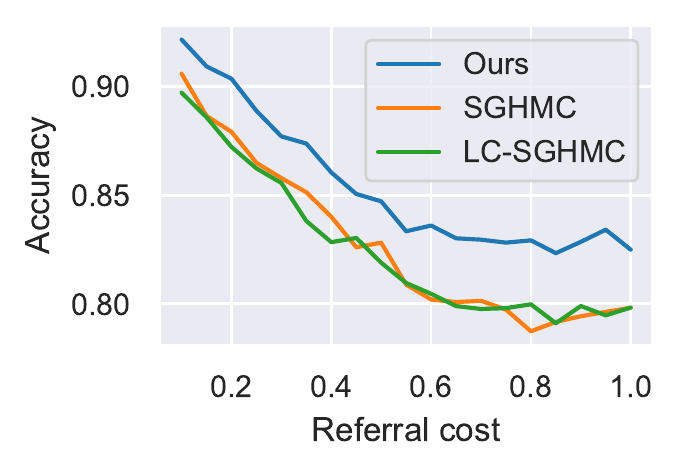}}
    {\includegraphics[width=0.245\textwidth]{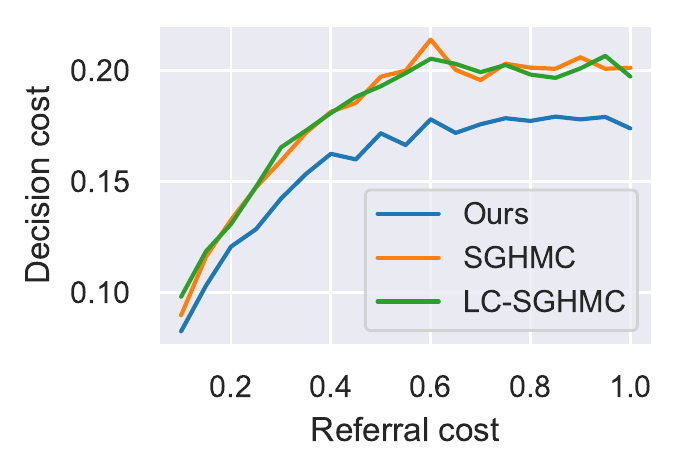}}
    {\includegraphics[width=0.245\textwidth]{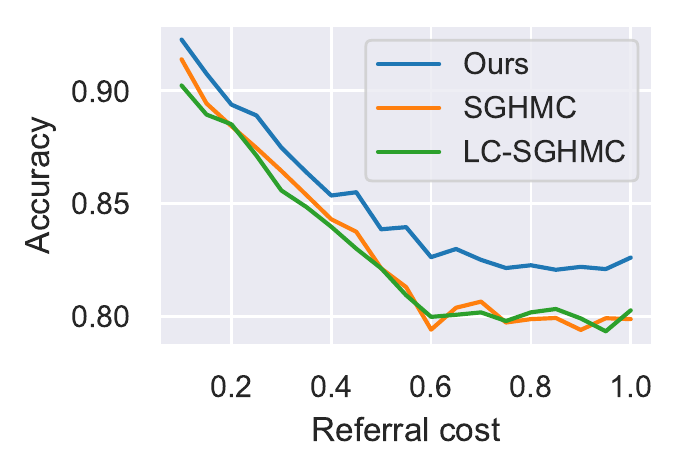}}
    \caption{\textbf{Selective classification}. Average test decision costs and test accuracy as a function of referral cost on CIFAR10 using a Bayesian ResNet18 model. The two left plots show results without using an amortized posterior predictive distribution, $S$, and the two plots on the right display results when using $S$. Accuracy is determined on those test data points that are not referred to the oracle. We note that introducing the amortized approximation $S$ does not adversely affect performance. Additional results involving negative log likelihood are presented in \cref{app:selective_decision_masking}.}
    \label{fig:selective_decision_making_results}
\end{figure*}

\begin{figure}
    \centering
    \includegraphics[width=0.8\linewidth]{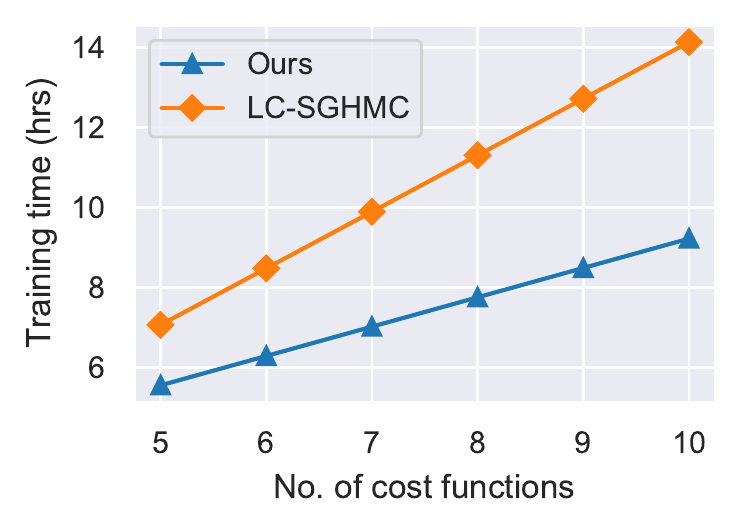}
    \caption{\textbf{Training cost comparison}. Wall-clock training time of our method and LC-SGHMC as a function of the number of decision cost functions.}
    \label{fig:selective_classification_training_time_comparison}
\end{figure}

Next, we consider the problem of selective classification, wherein the goal is to classify a data instance into one of $C$ classes or choose not to classify and instead refer it to an oracle. The corresponding decision problem thus involves selecting one of $C + 1$ decisions for each data instance. By adjusting the cost of a referral the decision making system can trade erroneous decisions for potentially expensive oracle feedback. Different users of such a system will likely prefer different trade-offs and as a result choose different referral costs. We however have no reason to believe that the different users would have different posterior beliefs. Since our method does not involve relearning the posterior beliefs when faced with changing cost functions, it is well suited for such selective classification problems. 

\textbf{Experimental setup: } We use the CIFAR10 \citep{krizhevsky2009learning} dataset along with SGHMC trained Bayesian ResNet18 \citep{He2016DeepRL} networks. To make the problem more challenging and encourage referrals, we contaminate the data via an additional data transformation. Under this contamination, we subject each image in the dataset by angle sampled uniformly at random from $[-30\degree, 30\degree]$. Next, following \citet{Murphy2012MachineL} (section 5.7.1.2), we define our selective classification decision cost function as, 
\begin{align*}
    \ell(c,y) = \left\{ \begin{array}{lr}
        0, & \text{for } y=c\\
        1, & \text{for } y \neq c\\
        r,  & \text{for } c=\text{referral}. 
        \end{array} \right.
\end{align*}
Here, $r$ denotes the cost of a referral. With this setup we examine a) the effectiveness of our method as a function of $r$, b) whether using the amortized posterior predictive distribution $S$ (maximizing \cref{eq:lcost_student_final}) adversely affects performance when compared to the non-amortized version (maximizing \cref{eq:lcost_final}), and c) the computational cost of learning post-hoc corrections under multiple decision-cost functions.  
Additional details around models, training procedures, hyperparameters, baselines, and cost function is given in the \cref{app:selective_decision_masking}. 

\textbf{Results:} The results for the experiment are presented in  \cref{fig:selective_decision_making_results}. As we would expect, lower values of the referral cost $r$ lead to more referrals and as a result, the models tend to make a decision only when they are very confident, leading to higher values of accuracy. Lower values of $r$ also result in lower average decision cost, as the models tend to refer to the oracle more, and due to the lower value of referral cost, the average decision cost reduces. As we can see from the comparison, the decision cost as well as accuracy of our post-hoc corrections outperforms those obtained using SGHMC and LC-SGHMC across all values of referral cost $r$. Furthermore, we observe that the accuracy and decision cost values are not adversely affected by using $S$ in place of the non-amortized posterior predictive distribution. We also compare the training time of LC-SGHMC and our post-hoc corrections as a function of the number of decision-cost functions.  The wall-clock times required by the two approaches are shown in \cref{fig:selective_classification_training_time_comparison}. All the experiments were run using the same GPU hardware (Nvidia Tesla V100) and under identical conditions for consistency in wall-clock time comparisons. We observe that our approach is significantly faster to train, and its computational cost grows at a slower rate with increasing number of cost functions. 

Finally, the fact that the amortized posterior predictive distribution $S$ does not adversely affect performance, is significant, in that it suggests that our approach would likely continue to scale with increasing network size, when storing and averaging over multiple Monte Carlo samples to compute the posterior predictive distribution is the bottleneck. Moving forward, we only consider experiments with the amortized posterior predictive $S$.  

\subsection{Decision making under poor data quality}
\begin{figure*}[h]
    \centering
    {\includegraphics[width=0.245\textwidth]{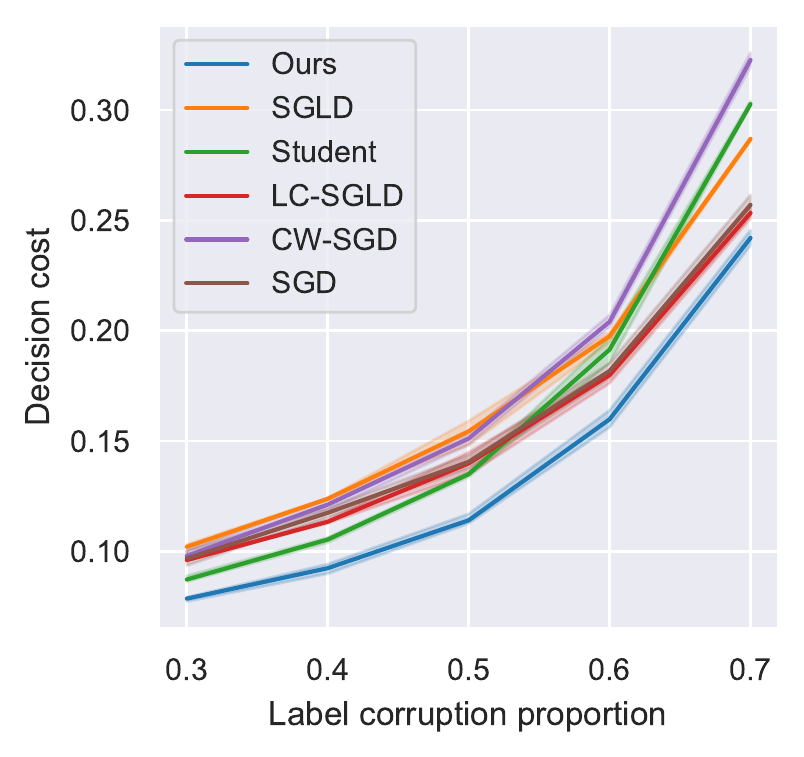}}
    {\includegraphics[width=0.245\textwidth]{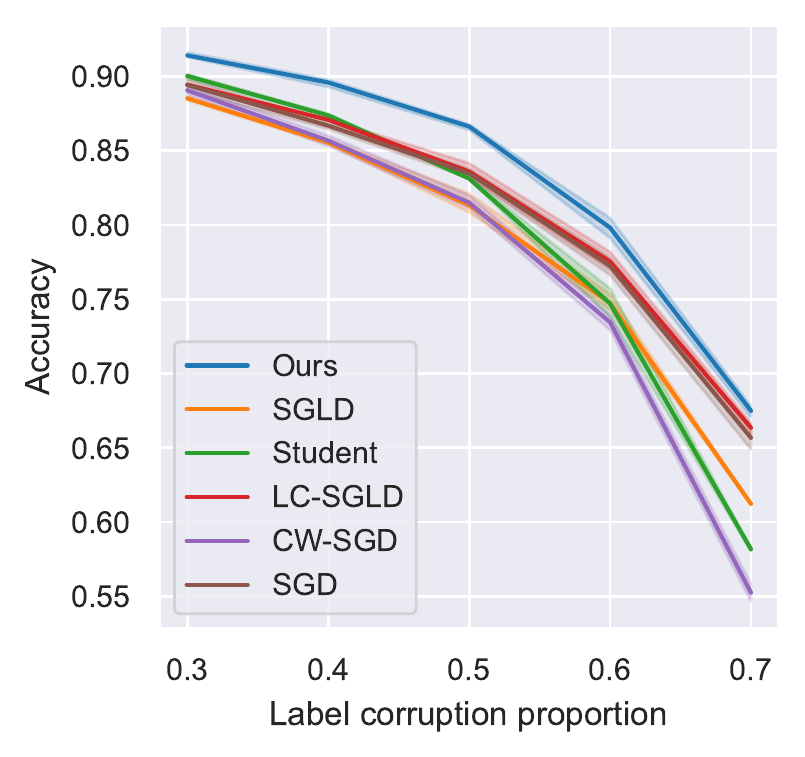}}
    {\includegraphics[width=0.245\textwidth]{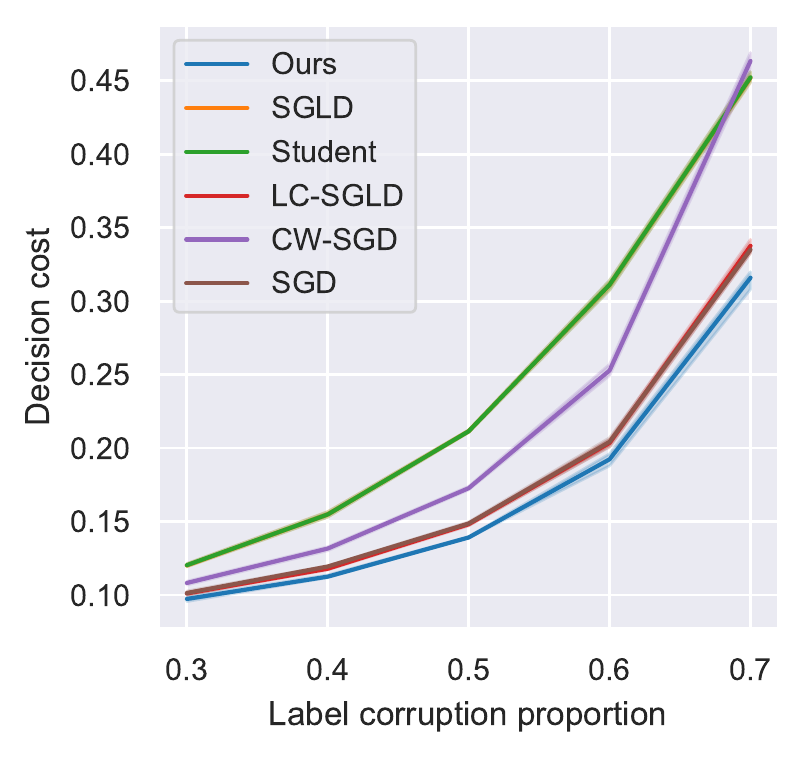}}
    {\includegraphics[width=0.245\textwidth]{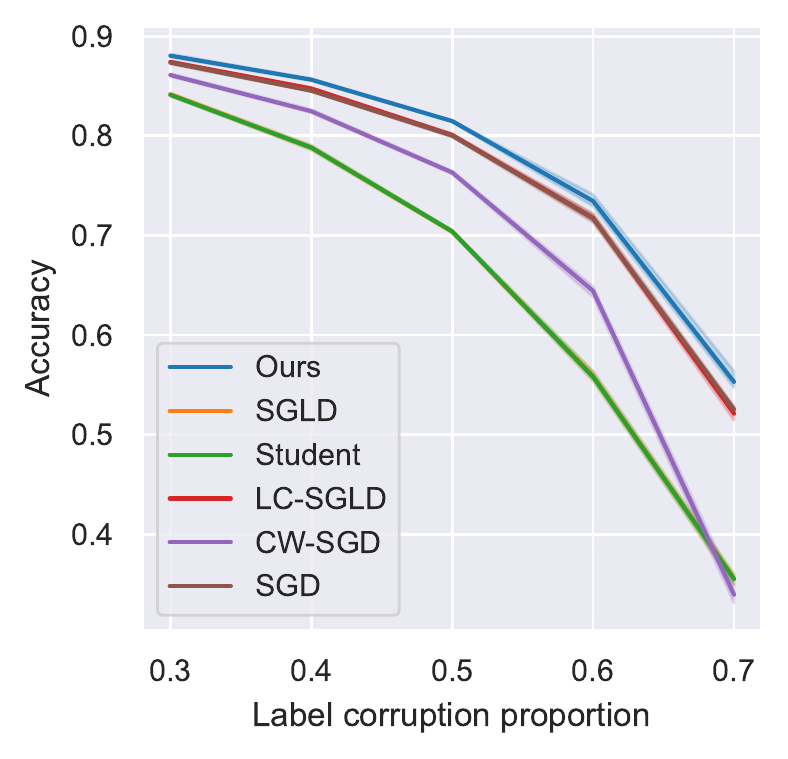}}
    \caption{\textbf{Decision-making under label corruption.} The two left plots illustrate performance comparison for different levels of label corruption on CIFAR10 using a Bayesian ResNet18 model. The right two plots illustrate performance comparison for different levels of label corruption on MNIST using an MLP with a single hidden layer of $200$ hidden units. The results are shown as mean $\pm$ std. dev. over 5 trials. Additional results involing negative log likelihood are presented in \cref{app:decision_making_poor_data_quality}.}
    \label{fig:label_corruption_results}
\end{figure*}

In the current era of big-data, it is not uncommon to have data sets of poor quality. In many settings, the data sets are labeled by crowdsourcing, as well as other automated techniques. These labeling techniques can often lead to noisy labels, and affecting downstream performance. Thus, it is important to understand how our method performs in this practical scenario. In practical scenarios, it is common to have asymmetric decision cost functions. This means that for making certain incorrect decisions, the cost can be higher or lower than the rest to encourage or discourage making those decisions. In this experiment, we also incorporate an asymmetric cost function similar to the one introduced earlier in the synthetic data experiments.

\textbf{Experimental setup: } We simulate label corruption on MNIST \citep{lecun1998mnist} and CIFAR10 \citep{krizhevsky2009learning}. For each dataset, we switch the true labels of a proportion of the training set to labels sampled uniformly at random. For MNIST, we use a simple multi-layer perceptron architecture with one hidden layer of 200 units, while for CIFAR10, we use the ResNet18 architecture \citep{He2016DeepRL}. It is worth noting that SGLD can be derived special case for the SGHMC (refer to \cref{app:SGHMC}). For each data set, we pick two classes to which we assign higher importance, and thus assigning a lower cost to the mistakes which involve choosing these classes as decisions. In MNIST, we assign a higher importance to classes 3 and 8, while in CIFAR10, we assign a higher importance to classes automobile and trucks. We use SGLD \citep{welling2011bayesian} for sampling from the posterior as well as the utility scaled posterior distribution. For a point-estimated model baseline, we introduce the class-weighted SGD (CW-SGD) baseline. In CW-SGD, we use our standard log loss on the neural network model, but assign a higher weight to the classes of interest. This encourages the model to make lesser mistakes on classes of higher importance. For additional details around models, training procedures, hyperparameters, and baselines, please refer to \cref{app:decision_making_poor_data_quality}. 

\textbf{Results: } We present the results of this experiment in  Figure \ref{fig:label_corruption_results}. For a comprehensive evaluation of performance, we vary the label corruption proportion between $0.3$ and $0.7$. For performance assessment, we look at the decision cost $(\downarrow)$, and accuracy $(\uparrow)$ on the standard test sets for each data set. 
While looking across the set of performance metrics, similar trends emerge. For lower levels of corruption, we notice that our post-hoc correction method performs similarly to LC-SGHMC and LC-SGLD, and marginally better than the uncorrected posterior predictive distribution and CW-SGD. However, as we increase the label corruption proportion to moderate levels, we observe that our method outperforms the baselines. Finally, with increasing corruption proportion, all methods are overwhelmed by the label noise, and we see a sharp dip in performance across all methods. We note that beyond achieving similar or lower decision costs, our approach also achieves higher accuracy and negative log-likelihoods than the competing methods. With these encouraging results in mind, we move towards our final experiment which looks at a real-world data set and demonstrates the benefits of amortized decision making.


\subsection{Semantic scene segmentation}
In this experiment, we consider the problem of semantic scene segmentation. Here, our goal is to segment an image into its components. This is achieved by labeling each pixel of an image with one of C known categories. Semantic segmentation can be useful for a variety of applications including aiding autonomous vehicles navigate the world. In such an application, it is crucial that the underlying decision problem of labeling pixels be solvable in near real-time. Existing loss-calibrated approaches struggle with such real-time requirements. through this experiment, we demonstrate both real time performance and improved decisions provided by our post-hoc loss correction framework.   

\begin{figure*}
    \centering
    {\includegraphics[width=0.9\linewidth]{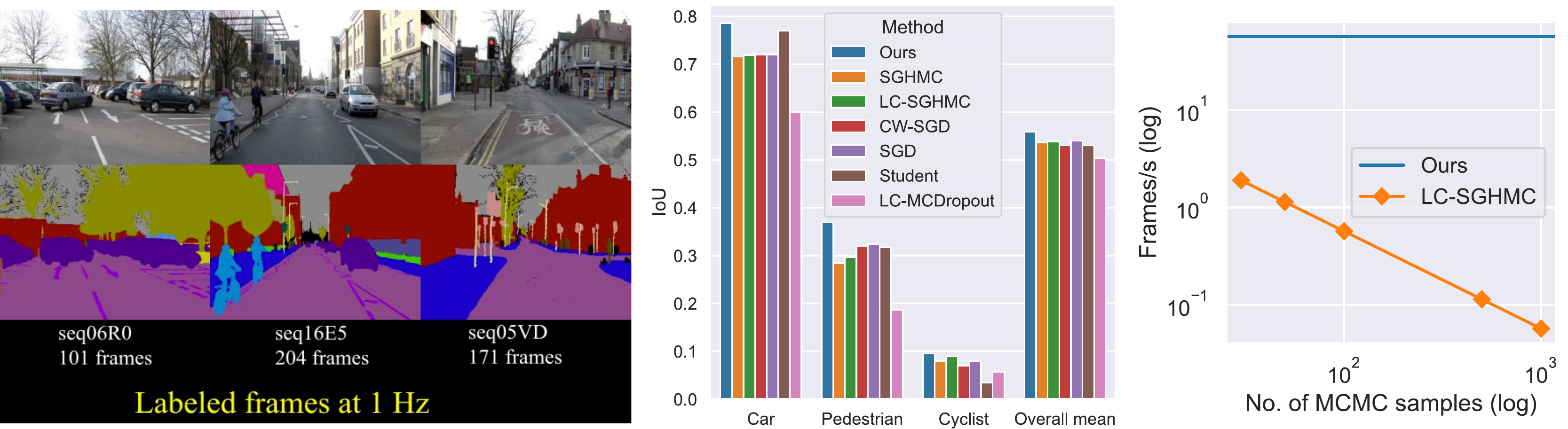}}
    \caption{ \textbf{Semantic Segmentation} (Left) Sample input images (top) and ground truth segmentations (bottom) from the CamVid dataset. (Middle) IoU scores acheived by different methods for classes deemed important by the cost function as well as overall mean IoU across all classes and all images. (Right) We compare the number of frames processed per second by our method and LC-SGHMC. Owing to amortization our approach is independent of the number of samples used to compute the Monte Carlo approximation to the posterior predictive distribution. In contrast, the number of frames processed per second by LC-SGHMC decreases dramatically (note the log scaling of the axes) with increasing number of samples.}
    \label{fig:segnet_camvid}
\end{figure*}

\textbf{Experimental setup: } For this experiment, we use the Camvid data set \citep{BrostowFC:PRL2008, BrostowSFC:ECCV08} which contains per-pixel labeled images captured using a camera on the dashboard of a car moving on the streets. An illustration of the data is provided in Figure \cref{fig:segnet_camvid}. The version of the data set that we use, contains a total of 12 class labels. Of these class labels, we assign a lower cost to false decisions that involve picking either of the pedestrian, cyclist or car class. This decision cost structure is inspired from the experiments of \citet{Cobb2018LossCalibratedAI}, as the goal of an autonomous car is to avoid these obstacles for safety reasons. By assigning a lower penalty to an incorrect decision of classifying a pixel to one of these three classes, we encourage our model to be more risk averse. For our model, we use the \emph{SegNet} model architecture \citep{Badrinarayanan2017SegNetAD}. SegNet is an auto-encoder style model which has previously been used for semantic scene segmentation. For additional comparison, we add the loss calibrated MC dropout baseline based on the algorithm presented in \citet{Cobb2018LossCalibratedAI}. We defer additional details around training procedures, hyperparameters, baselines and decision cost matrix to \cref{app:semantic_scene_segmentation}. 

\textbf{Results: } The aim of this experiment is two-fold. First, we want to evaluate the performance of our approach for the task, and compare it with relevant baselines. Secondly, we  want to assess the test-time efficiency of our approach against alternates. 

For assessing performance, we use the intersection over union (IoU) ($\uparrow$) metric that is commonly used to assess semantic segmentation performance. This metric evaluates the ratio of the area of overlap and the area of union while comparing the ground truth segmentation and model output segmentation for each class. In \cref{fig:segnet_camvid}, we present a performance comparison between our method and the baselines using the IoU metrics on test set. For the comparison, we look at each of the high utility classes separately, as well as we evaluate the overall mean IoU over all classes combined. We observe that our method performs consistently better on all the high utility classes, and thus doing a better job at capturing the preferences embedded in our cost function. Moreover, the mean IoU across all the classes indicates that our method does a better job overall for this task.

Next, we look at assessing test-time efficiency for our current task. The metric used for quantifying the test time efficiency is the number of frames that can be processed per second (FPS, $\uparrow$). For application such as autonomous driving, a slower processing pipeline can create a bottleneck when it comes to the efficacy for deployments. However, since we use a point-estimated model for our approach, it gives us inherent time savings when looking at test-time processing capabilities. In Figure \ref{fig:segnet_camvid}, we present a comparison of no. of frames/sec that can be processed between our approach and LC-SGHMC. Each frame (image) in the data set has a resolution of $360 \times 480$. We compute the time to process an image after loading both the image and model (or model ensemble for LC-SGHMC) on a Nvidia Tesla V100 GPU. While we expect monte carlo based approximations to perform much slower than our point-estimated model, it is increasingly evident looking at Figure \ref{fig:segnet_camvid} (middle) that the number of frames/s can be inhibitively low for practical applications, for even a smaller number of MCMC samples. While the performance for MCMC methods in Figure \ref{fig:segnet_camvid} is computed using 30 samples, we compute the frames/s metric over larger ensemble sizes to give a sense of why Monte Carlo integration at test time can be impractical.

To summarize, there are two key findings of this experiment. First, our approach leads to improved decision making by capturing class preferences better, and improves on the overall performance averaging across all twelve classes. Second, our method is better positioned for deployments requiring real-time performance as at decision time it requires a single forward pass and no Monte Carlo approximations.

\section{Discussion and future work}
In this paper we introduced a novel framework for post-hoc loss calibration of Bayesian neural networks for decision-making. Through comprehensive empirical evaluations ranging from synthetic data sets to practical applications involving real world data, we have demonstrated that our approach consistently produces lower cost, higher utility decisions than competing approaches. We also demonstrated that the framework by decoupling posterior inference from decision-making provides computational advantages at training time, and through amortization provides fast test-time decisions. Future directions include extensions to continuous decisions, more carefully exploring the effect of the choice of an inference algorithm on the quality of the correction, exploring post-hoc corrections under distribution shift, and studying the connections to generalized Bayesian inference~\citep{bissiri2016general, knoblauch2019generalized}. 


\subsubsection*{Acknowledgments}
Meet P. Vadera would like to thank the generous support of IBM Research during his internship for initiating and facilitating this work. Meet P. Vadera, and Benjamin M. Marlin were also partially supported by the US Army Research Laboratory under cooperative agreement W911NF-17-2-0196. The views and conclusions  contained  in  this  document  are  those  of  the  authors  and  should  not  be interpreted as representing the official policies,  either expressed  or  implied,  of  the  Army  Research  Laboratory  or  the  US  government. Finally, we would also like to thank Adam Cobb for helpful discussions through the course of this work.

\bibliography{references}

\clearpage
\appendix
\onecolumn
\section{Appendix}
\subsection{Post-hoc correction algorithm}
\label{app:algorithm}
We provide a high level overview of our method in \cref{alg:post_hoc_correction_algo_student}. Note that this algorithm assumes that we have access to a student model $S$ that can approximate the posterior predictive distribution $p(y|\mathbf{x}, \pazocal{D})$.

\begin{algorithm}[htbp]
\caption{Post-hoc loss correction algorithm}
\label{alg:post_hoc_correction_algo_student}
\begin{algorithmic}[1]
\Require Amortized posterior approximation $S(. | ., \omega)$, Calibration dataset $\pazocal{D'}$, minibatch size $B$, loss-calibrated model $q(.|., \lambda)$, number of training iterations $T$, initialization $\lambda_0$, supremum of the loss function $\ell$, $M$.
\State Initialize the loss calibrated model by setting its parameters to $\lambda_0$.
\For {t $\in [1, \ldots, T]$}
    \State Draw a mini-batch $\pazocal{B}\subset \pazocal{D'}$ of size $B$.
        \For{each data point $\mathbf{x}_b \in \pazocal{B}$}
        \State Compute $c_b =  \argmin_c \int_{y_b} \ell (c, y_b)q(y_b \mid \mathbf{x}_b, \lambda_t) dy_b$ 
        \EndFor
    \State Define $\displaystyle \pazocal{\tilde{L}}(\lambda, \mathbf{c}; \pazocal{B)} = - \sum_{b=1}^{B} \E_{q(y_b | \mathbf{x}_b, \lambda)}\left[ \frac{\ell(c_b, y_b)}{M}   \right] 
        - \textrm{KL}(q(y_b | \mathbf{x}_b, \lambda) || S(y_b | \mathbf{x}_b, \omega)))$
    \State Update $\lambda_{t+1} \leftarrow \text{SGDUpdate}(\lambda_{t}, \nabla_\lambda\pazocal{\tilde{L}}(\lambda, \mathbf{c}, \pazocal{B}))$ 
\EndFor
\State \Return Optimized parameters of the loss calibrated model, $\lambda_T$.
\end{algorithmic}
\end{algorithm}

\subsection{Derivation of the post-hoc correction objective}
\label{app:derivation}
We begin our derivation with the definition of log conditional gain as follows:
\small{
\begin{align*}
 & \log \pazocal{G}(\mathbf{h} = \mathbf{c} \mid \Dval; \lambda) \notag \\
 &=  \sum_{n=1}^{N} \log \int_{y} \bigg( u(h=c_n, y_n=y) q(y_n=y | \mathbf{x_n}, \lambda)  \notag \\
 &    \qquad  \qquad  \qquad \times \left. \frac{p(y_n=y | \mathbf{x_n}, \pazocal{D}, \theta^0)}{q(y_n=y | \mathbf{x_n}, \lambda)} d{y} \right) \notag \\ 
&= \sum_{n=1}^{N} \log \int_{y}  \bigg( q(y_n=y | \mathbf{x_n}, \lambda) \notag \\
& \left.  \qquad  \qquad  \qquad  \times \left(\frac{p(y_n=y | \mathbf{x_n}, \pazocal{D}, \theta^0) u(h=c_n, y_n=y)}{q(y_n=y | \mathbf{x_n}, \lambda)} \right)  \right) d{y} \notag \\
&= \sum_{n=1}^{N} \log \E_{q(y_n=y | \mathbf{x_n}, \lambda)} \left[\frac{p(y_n=y | \mathbf{x_n}, \pazocal{D}, \theta^0) u(h=c_n, y_n=y)}{q(y_n=y | \mathbf{x_n}, \lambda)} \right] \notag 
\end{align*}
}
Now, using Jensen's inequality, we obtain,
\begin{alignat}{4}
&& \log \pazocal{G}(\mathbf{h} = \mathbf{c} | \mathbf{X}) &\geq \sum_{n=1}^{N}  \E_{q(y_n | \mathbf{x_n}, \lambda)}\left[ \log \left(\frac{p(y_n | \mathbf{x_n}, \pazocal{D}, \theta^0) u(c_n, y_n)}{q(y)_n | \mathbf{x_n}, \lambda)} \right) \right] && \notag \\
&& &= \sum_{n=1}^{N} \E_{q(y_n | \mathbf{x_n}, \lambda)}\left[ \log u(c_n, y_n) \right] && \notag \\
&& & - \textrm{KL}(q(y_n | \mathbf{x_n}, \lambda) || p(y_n | \mathbf{x_n}, \pazocal{D}, \theta^0)) \notag \\
&& &\coloneqq \pazocal{U}(\lambda, \mathbf{c} ; \Dval) &&
\end{alignat}

\subsection{The variational gap}
Consider a single data instance $\mathbf{x}_n \in \Dval$. The gap between the log conditional gain and the lower bound then is,   
\begin{equation}
\small
\begin{split}
    \displaystyle \log \; &\pazocal{G}(h = c_n \mid \Dval; \lambda) - \pazocal{U}(\lambda, c_n, \Dval) =  \log \int u(h=c_n, y_n=y)p(y_n | \mathbf{x_n}, \pazocal{D}, \theta^0) dy - \pazocal{U}(\lambda, c_n, \Dval) \\
    & = \log \int u(h=c_n, y_n=y)p(y_n | \mathbf{x_n}, \pazocal{D}, \theta^0) dy - \int q(y_n=y | \mathbf{x}_n, \lambda) \log u(h=c_n, y_n=y)dy \\ 
    & + \int  q(y_n=y | \mathbf{x}_n, \lambda) \log  q(y_n=y | \mathbf{x}_n, \lambda) dy  - \int q(y_n=y | \mathbf{x}_n, \lambda) \log p(y_n | \mathbf{x_n}, \pazocal{D}, \theta^0) \\
    & = \log \int u(h=c_n, y_n=y)p(y_n | \mathbf{x_n}, \pazocal{D}, \theta^0) dy + \int q(y_n=y | \mathbf{x}_n, \lambda) \log \frac{q(y_n=y | \mathbf{x}_n, \lambda)}{p(y_n | \mathbf{x_n}, \pazocal{D}, \theta^0)u(h=c_n, y_n=y)} dy \\
    & = \log Z_n + \int q(y_n=y | \mathbf{x}_n, \lambda) \log \frac{q(y_n=y | \mathbf{x}_n, \lambda)}{p(y_n | \mathbf{x_n}, \pazocal{D}, \theta^0)u(h=c_n, y_n=y)} dy \\
    & = \int q(y_n=y | \mathbf{x}_n, \lambda) \log Z_n dy + \int q(y_n=y | \mathbf{x}_n, \lambda) \log \frac{q(y_n=y | \mathbf{x}_n, \lambda)}{p(y_n | \mathbf{x_n}, \pazocal{D}, \theta^0)u(h=c_n, y_n=y)} dy \\
    & = \int q(y_n=y | \mathbf{x}_n, \lambda) \log \frac{q(y_n=y | \mathbf{x}_n, \lambda) Z_n}{p(y_n | \mathbf{x_n}, \pazocal{D}, \theta^0)u(h=c_n, y_n=y)} dy \\
    & = \textrm{KL}\big[q(y_n | \mathbf{x}_n, \lambda) || \frac{p(y_n | \mathbf{x}_n, \pazocal{D}, \theta^0)u(h=c_n, y_n)}{Z_n} \big].
\end{split}
\end{equation}
\normalsize
Summing over all data points in $\Dval$ gives us \cref{eq:vgap}.
\subsection{Overview of Stochastic Gradient Hamiltonian Monte Carlo (SGHMC)}
\label{app:SGHMC}
Since we use SGHMC often as a part of this paper, we provide a brief overview of SGHMC in this appendix. SGHMC is a SGMCMC algorithm, which is a class of MCMC algorithms that function based on mini-batch gradients. This is particularly helpful for deep learning as traditional models and data sets are too large to compute the gradients of the likelihood on entire data, which can lead to a bottleneck in computation. SGHMC algorithm involves computing the $\tilde{U}(\theta)$ on a minibatch $\pazocal{B}$ as shown in the first equation, followed by running the next two update equations: 
\begin{align}
    \tilde{U}(\theta) &= - \log (p(\pazocal{B} | \theta)) - \log (p(\theta | \theta^0)) \\
    \theta_k &= \theta_{k-1} + v_{k-1}  \\
    v_{k} &= v_{k-1} - \alpha_k \nabla \Tilde{U} - \eta v_{k-1} + \sqrt{2(\eta - \hat{\gamma})\alpha_k} \epsilon_k 
\end{align}
In the above equations $k$ denotes the iteration, $v_k$ denotes the momentum term, $(1- \eta)$ denotes the momentum factor, $\eta_k$ is drawn from an identity gaussian distribution, $\nabla \Tilde{U}$ denotes the gradient approximation obtained using a minibatch, $\hat{\gamma}$ and $\alpha_k$ denotes the instantaneous step size. The process highlighted in the above equations is run iteratively to draw new samples from the posterior. For practical purposes, most implementations set $\hat{\gamma}$ to 0 \citep{Zhang2020Cyclical, Vadera2020URSABench}. Interestingly, SGLD can be derived from SGHMC by setting the momentum factor to 0. Our experiments use the SGHMC implementation provided by \cite{Vadera2020URSABench}.

\subsection{Additional experimental details and resuts: Synthetic Data Experiments}
\label{app:synthetic_data_experiments}
\textbf{Additional experimental Details:} For the SGHMC implementation in this experiment, we use a fixed learning rate of 0.1, a momentum of 0.5 and prior precision of 1.0. We run a burn-in phase of 300 iterations, and collect total 100 parameter samples with a thinning interval of 50 iterations. After this, we obtain the monte carlo approximation to the posterior predictive distribution $p(y| \mathbf{x}, \pazocal{D})$ on the additional training data points ($\pazocal{D'}$). For doing our post hoc correction, we use an MLP with 50 hidden units (as with the original model) and optimize the objective shown in \cref{eq:lcost_student_final} using Adam optimizer with a learning rate of 0.1 for 500 training iterations.

For BBVI implementation in this experiment, we set an identity gaussian distribution as our prior and maximize the evidence lower bound (ELBO) of our MLP-BNN using Adam optimizer with a learning rate of 0.01 over 5000 iterations. We use the local-reparameterization trick~\cite{kingma2015variational} to produce low variance stochastic gradients. Next, we again collect a total of 100 parameter samples and compute the Monte Carlo approximation of the posterior predictive distribution $p(y| \mathbf{x}, \pazocal{D})$ on the additional training data points ($\pazocal{D'}$). For doing our post hoc correction, we use an MLP with 50 hidden units (as with the original model) as our $q(.)$ model, and optimize the objective shown in \cref{eq:lcost_student_final} using Adam optimizer with a learning rate of 0.1 for 500 training iterations.

For KFAC-Laplace we perform a Laplace approximation about the maximum-a-posteriori (MAP) solution. As in standard Laplace approximation, the approximate posterior is represented by a Gaussian  centered at the MAP solution with its covariance set to the inverse of the Hessian.  We find the MAP solution by using Adam with a learning rate of 0.01 to maximize the negative log posterior.  Following \citet{ritter2018scalable} we use a block-diagonal, Kronecker-factored Hessian.

Finally, note that there are 100 training points in the original training set ($\pazocal{D}$), 500 additional unlabeled data points for our posterior correction ($\pazocal{D'}$), and 100 held out data points for testing from the same data distribution as $\pazocal{D}$.

\textbf{Additional experimental Details:} The full panel of results is given in Table \ref{tab:app:toy_data_results}.

\begin{table}[htbp]
\centering
\caption{\textbf{Results on synthetic data}. Test decision costs with and without post-hoc correction over $10$ replicates. Post-hoc correction consistently provides lower cost decisions. Results presented as mean $\pm$ std. dev.}
\label{tab:app:toy_data_results}
\begin{tabular}{ccccc}
\toprule
 & \begin{tabular}[c]{@{}c@{}}W/O post-hoc \\ correction\end{tabular} & \begin{tabular}[c]{@{}c@{}}W/ post-hoc \\ correction (ours)\end{tabular} & \begin{tabular}[c]{@{}c@{}} Avg. paired \\ diff. \end{tabular} & \begin{tabular}[c]{@{}c@{}} Std. (avg. paired \\ diff.) \end{tabular} \\ \midrule
VI       & 0.019 $\pm$ 0.011 & 0.016 $\pm$ 0.010 & 0.00146 & 0.001\\ \midrule
SGHMC    & 0.018 $\pm$ 0.008 &  0.017 $\pm$ 0.009 & 0.001 & 0.006 \\ \midrule
KFAC-Laplace & 0.021 $\pm$ 0.007 & 0.018 $\pm$ 0.008 & 0.002 & 0.005  \\
\bottomrule
\end{tabular}%
\end{table}

\subsection{Additional experimental details and results: Selective decision making}
\label{app:selective_decision_masking}
\textbf{Additional experimental Details:} In this experiment, we employ two methods for approximating the posterior predictive distribution $p(y| \mathbf{x}, \pazocal{D}, \theta^0)$: using a student model and pre-computing the posterior predictive distribution using the samples from SGHMC. For both cases, we generate the additional unlabeled training data $\pazocal{D'}$ by applying the same random transformation on original CIFAR10 data set, and generate 10 copies for every example by randomly rotating the image as described earlier. In the case where we use the student model, we distill the posterior predictive distribution using the approach of \citet{balan2015bayesian} and use the same $\pazocal{D'}$ for the distillation. Note that the approach by \citet{balan2015bayesian} allows us to interleave the sampling from $p(\theta| \pazocal{D, \theta^0})$ and distilling to a student model using an online approach. 

Finally, once we obtain some form of approximation for the posterior predictive distribution, we optimize either \cref{eq:lcost_student_final} or \cref{eq:lcost_final} depending on whether we have used the student or not. We use the same ResNet18 architecture for $q(.)$ model. For the SGHMC chains, we use a momentum of 0.7, a fixed learning rate of $10^{-3}$, and a prior precision of 5, and 1000 burn-in iterations (each iteration is a gradient step after a mini-batch). We run the SGMCMC sampling-distillation algorithm from \citet{balan2015bayesian} for a total of 100 epochs, and collect a total of 30 samples at the end of each of the last 30 epochs. The student model has the same ResNet18 architecture, and is trained using SGD with a one-cycle learning rate schedule \citep{smith2017cyclical}. The peak learning rate is 0.1 and the weight decay factor is $5 \times 10^{-4}$. We warm start our $q(.)$ model with the student and train it for 100 epochs using an SGD optimizer with a learning rate of $10^{-3}$. For the case where we don't use a student model, we train our $q(.)$ model using SGD with a cyclic learning rate schedule as it helps accelerate the training. The peak learning rate is 0.1 and the weight decay factor is . We use the equivalent number of training iterations as 100 epochs on the original data set $\pazocal{D}$. For sampling a mini-batch from $\pazocal{D'}$, we use sampling without replacement at every iteration. The mini-batch size for the entire experiment is set to 64. Also note that the SGHMC chain begins with a pretrained maximum-a-posteriori (MAP) solution. This pre-trained solution is obtained using an SGD optimizer with a one-cycle learning rate scheduler, with a peak learning rate of $0.05$, with the same prior precision, and is trained for 100 epochs using $\pazocal{D}$.

\textbf{Additional results: } In Figure \ref{fig:selective_decision_making_nll_results} we provide the NLL ($\downarrow$) results (on unrefered data points) from the selective decision making experiment.  
\begin{figure}[htbp]
    \centering
    {\includegraphics[width=0.4\textwidth]{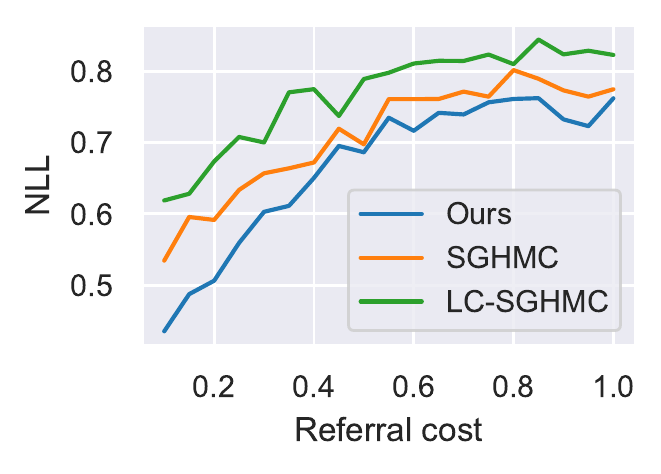}}
    {\includegraphics[width=0.4\textwidth]{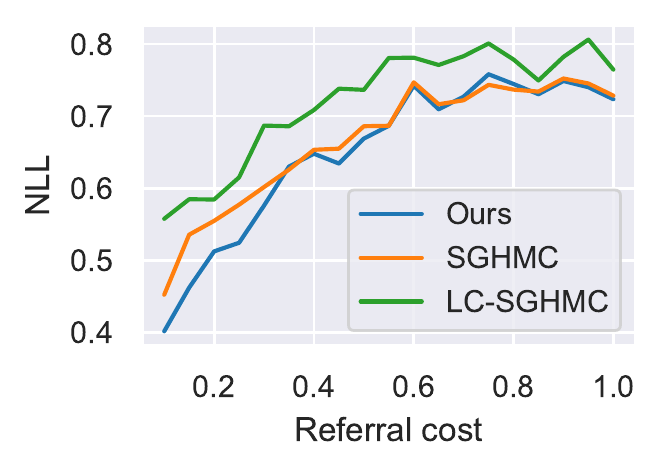}}
    \caption{\textbf{Selective classification}. NLL comparison on unreferred data points for different levels of label corruption for CIFAR10 data set using ResNet18 model. Top plot represent results from the experiment not using the student $S$ to approximate the posterior predictive distribution while the bottom plot represents results from the experiment using the student $S$ to approximate the posterior predictive distribution.}
    \label{fig:selective_decision_making_nll_results}
\end{figure}

\subsection{Additional experimental details and results: Decision making under poor data quality}
\label{app:decision_making_poor_data_quality}
\textbf{Additional experimental Details: } In this experiment, we use a ResNet18 model with CIFAR10 data set and a MLP with a single hidden layer of 200 units (MLP200) for MNIST. We generate $\pazocal{D'}$ online, by adding random pixel noise from a $\pazocal{N}(0, 0.05)$ distribution. Since we use student model, we follow the same algorithm and subsequent post-hoc correction as described in the previous subsection of the appendix. Student model  $S$ and our $q(.)$ model follow the same architecture as of the original model.

For ResNet18-CIFAR10 combination, we use SGLD with a fixed learning rate of $2\times10^{-5}$ and a prior precision of 50. The burn-in iterations is set to 10,000. We run the distillation algorithm to train the student model for 50 epochs, and collect 30 samples from the original model (also the teacher model here) from the end of the last 30 epochs. To train student model, we use SGD with a cyclic learning rate schedule. The maximum learning rate in the schedule is 0.05, with a momentum of 0.9 and a weight decay factor of $10^{-4}$. Once we obtain the student, we train our $q(.)$ model using the objective in \cref{eq:lcost_student_final}. For training the $q(.)$ model, we use SGD with a cyclic learning rate schedule. The maximum learning rate in the schedule is 0.05, with a momentum of 0.9. Note that for running SGHMC on ResNet18-CIFAR10 combination, we start with a pre-trained solution. This pre-trained MAP solution is obtained by running our original teacher model using the same data set $\pazocal{D}$. For the pre-training step, we use SGD with a cosine annealing learning with warm restarts. The initial learning rate in the schedule is 0.05, with a momentum of 0.9 and a weight decay factor of $5 \times 10^{-4}$. Each annealing phase is of 20 epochs, and we perform total 5 such phases.

For MNIST-MLP200 combination, we use SGLD (obtained by setting $\alpha=1$ in SGHMC) with a fixed learning rate of $10^{-4}$ and a prior precision of 6. The burn-in iterations is set to 10,000. We run the distillation algorithm to train the student model for 100 epochs, and collect 30 samples from the original model (also the teacher model here) from the end of the last 30 epochs. To train the student  model, we use SGD with a fixed learning rate of $10^{-3}$, with a momentum of 0.9 and a weight decay factor of $10^{-4}$. Once we obtain the student, we train our $q(.)$ model using the objective in \cref{eq:lcost_student_final}. For training the $q(.)$ model,we use SGD with a fixed learning rate of $10^{-3}$, with a momentum of 0.9. Note that for running SGLD on MNIST-MLP200 combination, we start with a pre-trained MAP solution as well. For the pre-training step, we use SGD with a cosine annealing learning with warm restarts. The initial learning rate in the schedule is 0.01, with a momentum of 0.9 and a weight decay factor of $1 \times 10^{-4}$. Each annealing phase is of 20 epochs, and we perform total 5 such phases.

The decision cost function for CIFAR10 is defined as shown below:
\begin{align}
\pazocal{\ell}(c, y) = 
\begin{cases}
0, & \text{for }  y=c, \\
0.7, & \text{for } y \neq c, \text{ and c} \in \{\text{automobile, truck}\} \\
1.0, & \text{otherwise}
\end{cases}
\end{align}

The decision cost function for MNIST is defined as shown below:
\begin{align}
\pazocal{\ell}(c, y) = 
\begin{cases}
0, & \text{for }  y=c, \\
0.7, & \text{for } y \neq c, \text{ and c} \in \{\text{3, 8}\} \\
1.0, & \text{otherwise}
\end{cases}
\end{align}

For the CW-SGD baseline, we assign a loss weight of 1.4 to the instance if the ground truth is either automobile or truck for CIFAR10 and if the ground truth either 3 or 8 for MNIST. The reason for this is that we want to our system to assign higher importance to these classes, and thus a mistake of making an incorrect decision with these ground truth classes should incur higher loss. Similarly, the decision cost function highlights the same fact as it places a lesser penalty if our decision system predicts these important classes.

\textbf{Additional results: } In Figure \ref{fig:selective_decision_making_nll_results} we provide the NLL ($\downarrow$) comparison from the experiment on decision making under poor data quality.  
\begin{figure}[htbp]
    \centering
    {\includegraphics[width=0.4\textwidth]{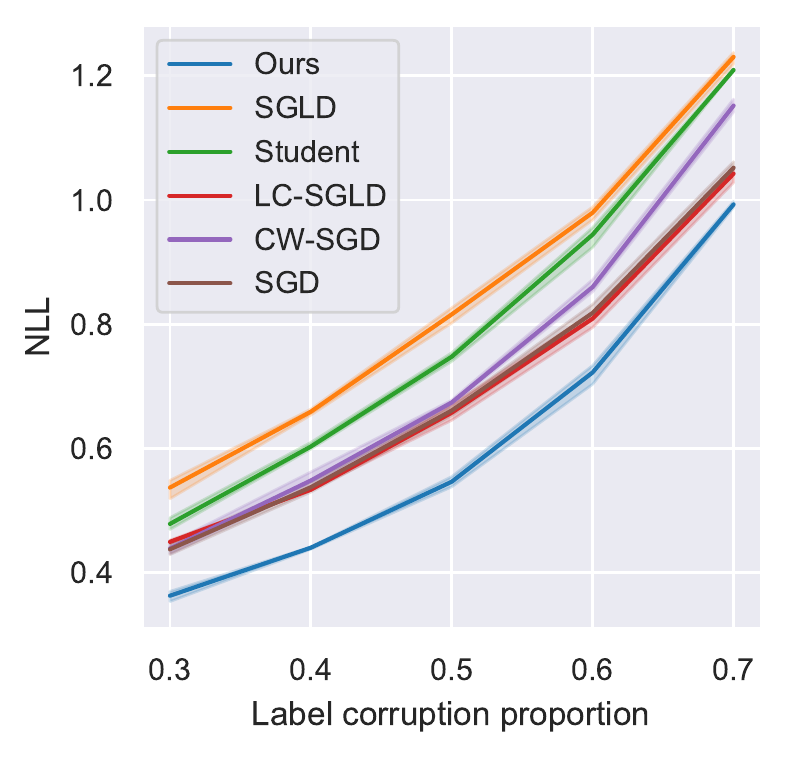}}
    {\includegraphics[width=0.4\textwidth]{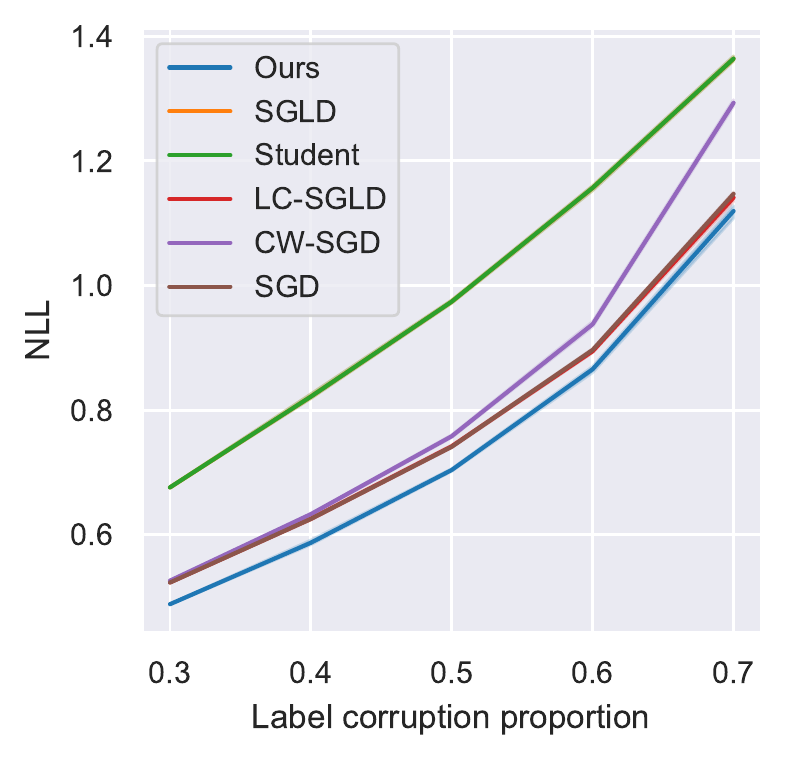}}
    \caption{\textbf{Decision-making under label corruption.} (Top) Performance comparison for different levels of label corruption for CIFAR10 data set using ResNet18 model  and (Bottom) MNIST data set using an MLP with a single hidden layer of 200 hidden units. The results are shown as mean $\pm$ std. dev. over 5 trials.}
    \label{fig:selective_decision_making_nll_results}
\end{figure}

\begin{table*}[htbp]
\begin{center}
\caption{\textbf{Semantic scene segmentation.} Decision cost matrix for the semantic scene segmentation experiment.}
\label{tab:segnet_cost_matrix}
\noindent
\begin{tabular}{cc*{14}{c}}
\multicolumn{1}{c}{} &\multicolumn{1}{c}{} &\multicolumn{10}{c}{\textbf{Decision}} \\ 
 & 
\multicolumn{1}{c}{\textbf{Cost}} & 
\multicolumn{1}{c}{\small Sk.} & 
\multicolumn{1}{c}{\small Bu.} & 
\multicolumn{1}{c}{\small Po.} & 
\multicolumn{1}{c}{\small Ro.} & 
\multicolumn{1}{c}{\small Pa.} & 
\multicolumn{1}{c}{\small Tr.} & 
\multicolumn{1}{c}{\small Si.} & 
\multicolumn{1}{c}{\small Fe.} & 
\multicolumn{1}{c}{\small Ca.} &
\multicolumn{1}{c}{\small Pe.} &
\multicolumn{1}{c}{\small Cy.} &
\multicolumn{1}{c}{\small Un.} \\ 
\multirow{10}{*}{\rotatebox{90}{\textbf{Ground Truth/Prediction}}} 
&\small Sky             & 0.  & 0.8 & 0.6 & 0.6 & 0.6 & 0.6 & 0.6 & 0.6 & 0.4 & 0.4 & 0.4 & 0.6  \\ 
&\small Building        & 0.8 & 0.  & 0.6 & 0.6 & 0.6 & 0.6 & 0.6 & 0.6 & 0.4 & 0.4 & 0.4 & 0.6  \\ 
&\small Pole            & 0.8 & 0.8 & 0.  & 0.6 & 0.6 & 0.6 & 0.6 & 0.6 & 0.4 & 0.4 & 0.4 & 0.6  \\ 
&\small Road            & 0.8 & 0.8 & 0.6 & 0.  & 0.6 & 0.6 & 0.6 & 0.6 & 0.4 & 0.4 & 0.4 & 0.6  \\ 
&\small Pavement        & 0.8 & 0.8 & 0.6 & 0.6 & 0.  & 0.6 & 0.6 & 0.6 & 0.4 & 0.4 & 0.4 & 0.6  \\ 
&\small Tree            & 0.8 & 0.8 & 0.6 & 0.6 & 0.6 & 0.  & 0.6 & 0.6 & 0.4 & 0.4 & 0.4 & 0.6  \\ 
&\small Sign            & 0.8 & 0.8 & 0.6 & 0.6 & 0.6 & 0.6 & 0.  & 0.6 & 0.4 & 0.4 & 0.4 & 0.6  \\ 
&\small Fence           & 0.8 & 0.8 & 0.6 & 0.6 & 0.6 & 0.6 & 0.6 & 0.  & 0.4 & 0.4 & 0.4 & 0.6  \\ 
&\small Car             & 0.8 & 0.8 & 0.6 & 0.6 & 0.6 & 0.6 & 0.6 & 0.6 & 0.  & 0.4 & 0.4 & 0.6  \\ 
&\small Pedestrian      & 0.8 & 0.8 & 0.6 & 0.6 & 0.6 & 0.6 & 0.6 & 0.6 & 0.4 & 0.  & 0.2 & 0.6  \\ 
&\small Cyclist         & 0.8 & 0.8 & 0.6 & 0.6 & 0.6 & 0.6 & 0.6 & 0.6 & 0.4 & 0.2 & 0.  & 0.6  \\ 
&\small Unlabelled      & 0.8 & 0.8 & 0.6 & 0.6 & 0.6 & 0.6 & 0.6 & 0.6 & 0.4 & 0.4 & 0.4 & 0.  \\ 
\end{tabular}
\end{center}
\end{table*}

\subsection{Additional experimental details: Semantic Scene Segmentation}
\label{app:semantic_scene_segmentation}
For the SGHMC implementation in this experiment, we use an initial learning rate of 0.01, a momentum of 0.5 and effective weight decay of $5\times10^{-4}$. We also use a cosine annealing learning rate scheduler which decays the learning rate to 0. We run a burn-in phase of 1000 iterations, and collect total 30 parameter samples with a thinning interval of 500 iterations. 
We run the distillation algorithm alongside the sampling to train student model for a total of 15000 iterations after the burn-in. To train student model, we use SGD with a cyclic learning rate schedule. The maximum learning rate in the schedule is 0.15, with a momentum of 0.9 and a weight decay factor of $10^{-4}$. Once we obtain the student, we train our $q(.)$ model using the objective in \cref{eq:lcost_student_final}. For training the $q(.)$ model,we use SGD with a cyclic learning rate schedule. The maximum learning rate in the schedule is 0.05, with a momentum of 0.9. Note that for running SGHMC we start with a pre-trained solution. The pre-trained solution used in this experiment is the CW-SGD solution. More details on CW-SGD solution is given later in this section. Once we obtain the student model $S$, we train the $q(.)$ model, using SGD with a cyclic learning rate schedule. The maximum learning rate in the schedule is 0.05, with a momentum of 0.9 for a total of 15000 iterations.

For the CW-SGD baseline, we use a cyclic learning rate schedule and train the Segnet model for 15000 iterations with a maximum learning rate of 0.05, weight decay of $10^{-4}$ and momentum of 0.9. The classes of importance here are the car, cyclist and pedestrian, and hence the loss weights for these classes are set to 1.4 while rest of the loss weights are set to 1.

For the loss calibrated MC dropout baseline, we use the pretrained SGD solution, and fine-tune it with Adam optimizer using a learning rate of $10^{-4}$ for 1000 training iterations. The dropout strength is 0.3 and is added in the decoding phase of the model. For computing the posterior predictive distribution at train, we use 5 forward passes, and during testing, we use 30 forward passes. Additional details about this algorithm can be found in \cite{Cobb2018LossCalibratedAI}

The decision cost matrix for this experiment is shown in \cref{tab:segnet_cost_matrix}. This decision cost matrix puts less penalty on decisions involving the 
\end{document}